# Face Verification and Forgery Detection for Ophthalmic Surgery Images

Kaushal Bhogale
Department of Information Technology
Vivekanand Education Society's Institute of Technology
Mumbai, India
2015kaushal.bhogale@ves.ac.in

Nishant Shankar
Department of Computer Engineering
Vivekanand Education Society's Institute of Technology
Mumbai, India
2015nishant.shankar@ves.ac.in

Adheesh Juvekar
Department of Information Technology
Vivekanand Education Society's Institute of Technology
Mumbai, India
2015adheesh.juvekar@ves.ac.in

Asutosh Padhi
Department of Information Technology
Vivekanand Education Society's Institute of Technology
Mumbai, India
2015asutosh.padhi@ves.ac.in

*Abstract*—Although modern face verification systems are accessible and accurate, they are not always robust to pose variance and occlusions. Moreover, accurate models require a large amount of data to train. We structure our experiments to operate on small amounts of data obtained from an NGO that funds ophthalmic surgeries. We set up our face verification task as that of verifying pre-operation and post-operation images of a patient that undergoes ophthalmic surgery, and as such the post-operation images have occlusions like an eye patch. In this paper, we present a system that performs the face verification task using one-shot learning. To this end, our paper uses deep convolutional networks and compares different model architectures and loss functions. Our best model achieves 85% test accuracy. During inference time, we also attempt to detect image forgeries in addition to performing face verification. To achieve this, we use Error Level Analysis. Finally, we propose an inference pipeline that demonstrates how these techniques can be used to implement an automated face verification and forgery detection system.

I. INTRODUCTION

With the increase in the effectiveness of intelligent automated systems, organizations are investing heavily in face recognition to bolster automated identity verification. Albeit face recognition is a relatively accessible technology today, the creation of automated systems that are resilient to occlusions and capable of handling a variety of lighting conditions still poses a challenge. Apart from these deployment stage considerations, it is found that feeding more data to a deep neural network during training results in a more accurate model. In production scenarios, organizations without sufficient technological infrastructure cannot leverage accurate face recognition economically.

In our paper, we approach this problem of designing deep neural networks for face verification that are agnostic to environmental conditions like lighting or occlusions and can be trained with a single training example per individual. Though the design of our system is generic enough to be applicable over many contexts, we evaluate our system on real-world data from an NGO. Like many organizations that are seeking to utilize face recognition in their workflow, NGOs are hard-pressed in terms of time and workforce. We consider a scenario where the NGO funds eye surgeries after verifying the identity of the applicant.

Identity verification can often create bottlenecks in the process of releasing funds to patients, especially if the NGO is understaffed. In the current ecosystem, as shown in Fig. 1, the NGO requests eye hospitals to send in pre-operation (pre-op) and post-operation (post-op) photographs of the patient's face for each surgery. These photographs are scrutinized by NGO workers, and patients are granted funding if the photographs are found to be legitimate.

Since the number of images the NGO receives is large, it becomes tedious to cross-verify all of them manually. An automated system that assists in authentication is required. It is also possible that multiple images of the same person are sent to the NGO, and these cases need to be detected and relayed to the NGO. Moreover, fraudulent and photoshopped images can be sent to the NGO, which also need to be identified.

The images of the patients under consideration are captured by the hospital and therefore can vary from clinic to clinic. Our paper aims to address this unconstrained face verification task, and also tackle partial occlusions in the images. Facial recognition under partial occlusion is certainly challenging, but achieving high accuracy with limited data is a more pressing challenge.

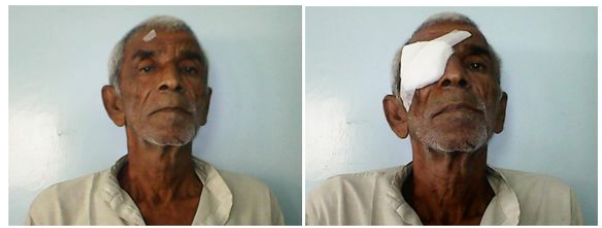

Fig. 1. A typical image pair from our dataset.. The image on the left is pre-op image and the image on the right is the post-op image, in which the patient wears an eye patch that occludes the operated eye.

Instead of relying on black box algorithms that consume huge amounts of data, our paper describes the design of intelligent algorithms that learn to recognize faces with the eye covered even when not much data is available.

## II. Background and Related Work

There has been considerable research in face recognition and verification, using both neural networks and a feature engineering-based approach. Since the breadth of this research is large, we discuss works relevant to the problems of pose variance, partial occlusions and limited availability of data.

Li et al. have worked on facial recognition using HOG and PCA algorithm[6]. They use a Haar classifier to detect faces in the image. HOG descriptors are found for the identified facial regions, and then a dimensionality reduction technique like PCA is applied[12]. Using these extracted features, an SVM classifier is used to recognize faces[11]. This approach is however not designed to handle occlusions in the images.

Xavier P. Burgos-Artizzu et al.[5] introduced a new method for facial verification of occluded images. It is an extension of CPR[10] and improves upon it by making CPR robust to occlusions. They use facial landmark detection on datasets like LFPW, LFW, and HELEN. RCPR uses annotated data as the ground truth value during training. However, due to variance in the nature of images, the pre-trained RCPR model performs poorly while predicting landmarks on our dataset. Employing RCPR on our dataset would require annotating data manually, which is an exercise that goes against our philosophy of making the training process more convenient.

Schroff et al.[1] describe an algorithm called FaceNet, which used a deep convolutional network to convert images to Euclidean spaces where distances directly correspond to a measure of face similarity. It is trained using a triplet loss function. The pre-trained model, while having a high accuracy on LFW, was not suitable for our dataset. Moreover, FaceNet required extraction of the face from the image in the initial stage which was not accurate due to the presence of occlusions like an eye patch.

The problem of training deep neural networks on limited data has been addressed in [13] using siamese neural networks for the task of character recognition. In [14], on the other hand, a face representation is learned. They take the concept of training on limited data (termed as the underrepresented set) further, by evaluating against a base set trained on a larger number of samples. They modify the loss function by adding a loss term specifically for underrepresented sets. These approaches are ways to perform "one-shot recognition", training on only one or a limited number of samples per individual.

Before performing face verification on our images, we need to identify whether the images are photoshopped or not. We discuss research pertaining to traditional photoshop detection. One such method to detect image forgery uses EXIF data [8]. When an image is captured using a camera, metadata like date, time, camera model, geolocation, etc., are saved concomitantly. In some cases, the metadata might also have information about the software that is used to edit or manipulate the image. However, this method is highly unreliable as many images do not have EXIF metadata.

## III. Dataset

Our dataset contains photographs of patients who underwent an eye surgery. These photographs were captured in a hospital environment, using a digital camera. The images are in standard JPEG format having dimensions 640x480. The dataset is divided into two sets, pre-operation images and post-operation images. There exists a single pair of pre-operation and post-operation images for each face in the dataset. Overall, the dataset has 1000 such pairs. The post-operation pictures have a part of the patient's face covered with a patch, as a result of the surgery. Additionally, faces have pose variations and occlusions other than the eye patch, as shown in Fig 2. Another notable challenge in our dataset is that there is only a single training example for each individual.

## IV. Preprocessing

Ideally, any face verification pipeline requires detecting the face from the image, and extracting the region of interest. This makes the models more accurate in downstream applications. We try two approaches, namely the Haar Classifier and Faster-RCNN[9]. Since face detection models rely on facial features, which can often be occluded, they perform poorly on post-operation images. Due to this, face detection and cropping yielded unsatisfactory results on our dataset.

If the image is passed onto further stages of the pipeline as it is, the noise it contains in the form of the background might hinder the learning procedure. To avoid this, the first stage of preprocessing involves background removal. To achieve this, we perform Canny Edge Detection[7] and dilate it using a square support uniform weight filter. This marks the boundaries of the face. We flood fill the edges to create a mask for the background. Using this mask we delineate the foreground from the rest of the image.

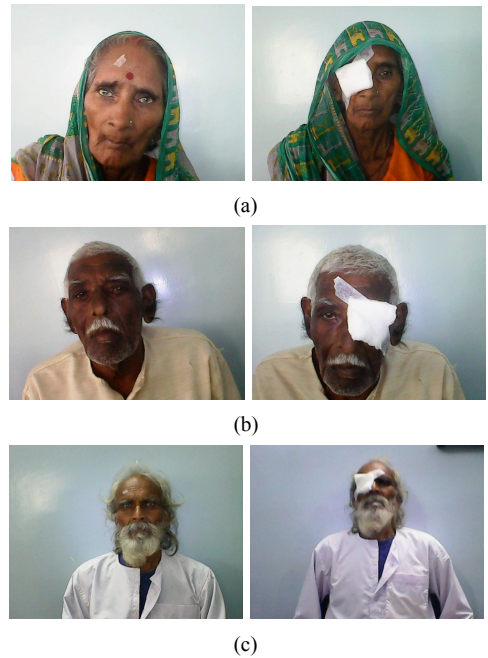

Fig. 2. Some example images from the dataset. (a) has occlusion other than eye patch (in this case, clothes), (b) and (c) show pose variation

Our dataset has only one image pair per person. To obtain higher accuracy, we need to increase the number of images per person. This is done using data augmentation[4]. A way of thinking about this augmentation is to consider that the changes in the proportion and scale of the image account for variations in posture. We perform the following affine transformations on the image with random probability according to the parameters given in Table I.

TABLE I. Parameters used for affine data transformations

| | |
|---|---|
| **Flip-left-right** | (probability=0.5) |
| **Rotate** | (probability=0.9, max-left-rotation=20, max-right-rotation=20) |
| **Zoom** | (probability=0.3, min-factor=1, max-factor=1.3) |
| **Random-distort** | (probability=0.6, grid-width=4, grid-height=4, magnitude=1) |

## V. Models

The goal of designing models for face verification is to map an input image $x$ to a feature space $R^d$ where $f(x)$ is the embedding of $x$. This embedding is unique for every face encountered by our system. To achieve the creation of this embedding, which can later be used to calculate vector norms, we explore different model architectures. In our case, the input images can be thought of in two ways - on the basis of the identity of the patient in the image, or the operation status (pre-op or post-op) of the individual. We train different siamese networks for the latter consideration (PRE-PRE, POST-POST and PRE-POST networks), and the loss function in each network operates on a low-dimensional representation of the patient's identity. $i$ represents a selection of a triplet out of a mini batch of size $N$.

### A. Siamese Network : Contrastive Loss

We define our siamese network similarly to [16]. Siamese networks consist of two identical convolutional networks that operate on an image pair $(x_u, x_v)$. The parallel CNNs have shared weights. In (1), the contrastive loss function we use to train each siamese network minimises the L2 square norm between the embeddings of the same individual ($x_u$ and $x_v$ are images of the same person), and maximizes this distance for embeddings of different people ($x_u$ and $x_v$ are images of different people). $D_w$ represents the L2 norm $\|f(x_u) - f(x_v)\|_2^2$ and Y represents the label associated with the training step.

$$L = \sum_i^N \frac{1}{2}[(1-Y)D_w + (Y)\{max(0, m - D_w)\}] \quad (1)$$

### B. Triplet Loss : Good things come in three

The triplet loss works on an input triplet instead of pair, and infuses the philosophy of the contrastive loss right at the optimization step itself as shown in (2). Here, as [2] notes, the triplet loss contextualizes the similar and dissimilar inputs during optimization, instead of deriving these pairwise losses independently. For each type of the aforementioned three networks, we sample a triplet from the dataset such that there is a sufficient mix of common and hard examples. This enables our model to learn an effective embedding space for all individuals. The input triplet can be defined as ($x_i^a$, $x_i^p$, $x_i^n$) where $x_i^a$ is the anchor image, $x_i^p$ is a positive sample and $x_i^n$ is a negative sample. α represents the margin between positive and negative pairs.

$$L = \sum_i^N \left[ \|f(x_i^a) - f(x_i^p)\|_2^2 - \|f(x_i^a) - f(x_i^n)\|_2^2 + \alpha \right]_+ \quad (2)$$

## VI. Experiments

### A. Evaluation

We evaluate the effectiveness of the discussed loss functions in Table II. Each model has been trained on pre-post images. Apart from the accuracy, which is highest in the case of the triplet loss, we also measure the false acceptance and rejection. By false acceptance, we mean that our system identifies two different individuals to be the same person. A false rejection entails identification of two images of the same person as belonging to different people From Table II, we can say that triplet loss is a better representation learning technique than contrastive loss. We plot the loss curve of this model in Fig. 3.

### B. Does the image background matter?

To verify whether the process of background removal is justified, we compare our best model obtained so far with a similarly designed model that is trained on images with the background preserved. From Table II, it is clear that the removal of background during the preprocessing step is justified, and it improves the accuracy of our model.

TABLE II. Metrics for different Loss Functions

| Loss Function | False Acceptance | False Rejection | Accuracy |
|---|---|---|---|
| Contrastive loss (with background removal) | 17.6% | 30% | 76.2% |
| Triplet loss (No background removal) | 18.8% | 67.2% | 57% |
| Triplet loss (with Background removal) | 0.4% | 29.6% | 85% |

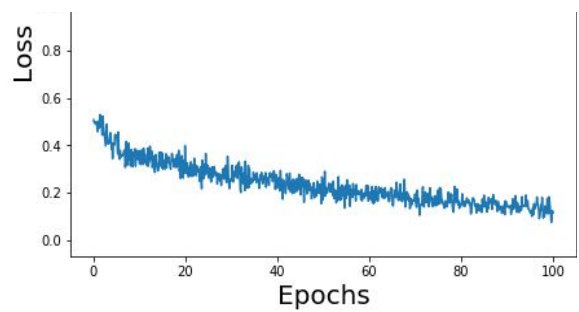

Fig. 3. Loss plot for 100 epochs for PRE-POST model using triplet loss as the loss function.

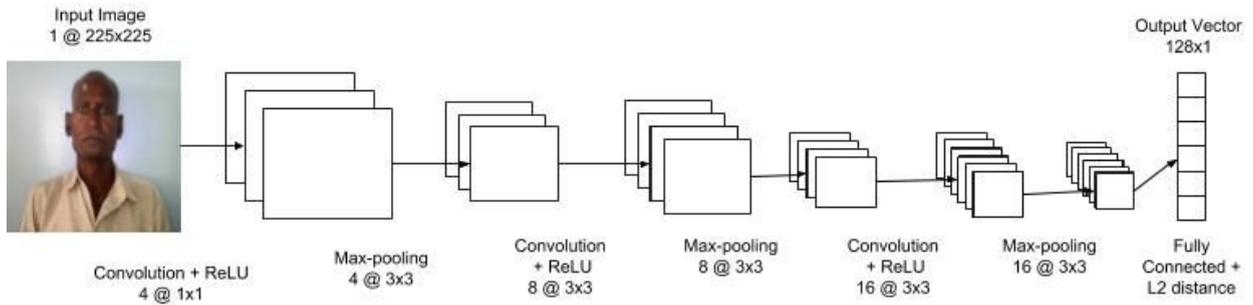

Fig. 4. All the models we have trained are based on this single CNN architecture. We use alternate convolution and Pooling layers with a ReLu activation function. The Siamese neural network trained with a contrastive loss function has two identical CNNs with shared weight matrices. The networks join at the fully connected layer. The network trained with the triplet loss function has three identical CNNs (for the input triplet), and the respective shared weights join at the fully connected layer.

### C. Generalizability of a Pre-post network

To evaluate whether specialized networks for the three different tasks are necessary, we test our Pre-post model on different image subsets. From Table III, though the Pre-post model performs well in the duplicate (pre-pre and post-post) identification task, specialized networks for those tasks surpass the Pre-post model.

TABLE III. TESTING MODELS ON DIFFERENT TYPES OF IMAGE DATA

| Image Description | Tested On | Accuracy |
|---|---|---|
| Pre-pre comparisons with background removal | PRE-PRE model | 98.6% |
| Pre-pre comparisons with background removal | PRE-POST model | 94.6% |
| Post-post comparisons with background removal | POST-POST model | 98% |
| Post-post comparisons with background removal | PRE-POST model | 92.4% |

### VII. DETECTING IMAGE FORGERY

Our paper defines a fraudulent image as an image that has been retouched or manipulated using an image editing software or tool. To detect these forgeries, we utilize a preexisting approach called Error Level Analysis[3].

This algorithm takes advantage of the fact that JPEG compresses its image every time it is changed. For JPEG compression, the image is divided into 8x8 matrices. Each matrix has an error level artifact associated with it which is generated when it is saved (compressed). In a genuine image, each matrix will have more or less the same error level artifact as they have been compressed the same number of times. But when parts of the image are hampered, their matrices have a different Error Level Attribute as compared to the rest of the image as it has been compressed a different number of times.

So when an image is to be scrutinized, it is recompressed and the difference between the 2 images (the forged image and the recompressed image) is taken which then highlights the exact position of the forgery.

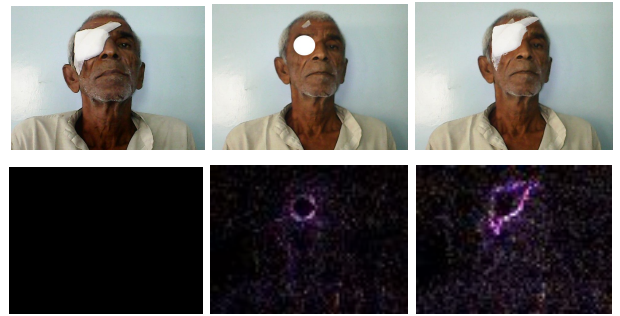

Fig. 5. Images on the bottom row show the corresponding ELAs of the images on the top row. Image on bottom left shows pitch black color as it is the ELA of a genuine image. Rest have been highlighted exactly where they were tampered.

This happens because the matrices of the same image have been compressed a different number of times.

### VIII. INFERENCE PIPELINE

Now that we have trained our model, we describe a pipeline to draw inferences on the given pair of pre-op and post-op images. The inference phase comprises of detecting the forgery, and performing face verification by looking up an embedding database. We compare this embedding with the existing embedding database and if the L2 norm between these embeddings is below the specified threshold $\theta$, our system verifies the identity of the individual.

**Algorithm 1 : Fraud Detection and Face verification**

**Initialisation:**
1: Load PRE-POST, PRE-PRE and POST-POST trained models.
2: Define the threshold : $\theta$

**Pre-op and Post-op image comparison:**
Given image-pair(*pre*, *post*):
1: CHECK-IMAGE-FORGERY(*pre*).
2: CHECK-IMAGE-FORGERY(*post*).
3: If any of the image is forged, display the ELA and RETURN.
4: **Generate the embeddings for the images by a forward pass on the PRE-POST model.**
   *emb*1, *emb*2 ← PRE-POST(*pre*, *post*)
5: *if* L2-DISTANCE(*emb*1, *emb*2) ≤ $\theta$ :
   a. Accept image pair and RETURN.
6: Else:
   a. Reject the image pair.

**Pre-op Duplicate Verification (Similar for Post-op):**
Given pre-image( *pre* ):
1: **Generate the embeddings by a forward pass on the PRE-PRE model.**
   $emb \leftarrow$ PRE-PRE( *pre* )
2: For-each $emb_{prev}$ in the database:
   a. *if* L2-DISTANCE( *emb*, $emb_{prev}$ ) $\leq \theta$ :
      i. Report *pre* as duplicate to the image in the database.
   b. Store *emb* in a database.

## IX. FUTURE SCOPE

In taking our work further, we have identified two primary areas in which our research can be expanded. Firstly, the performance of our model heavily depends on the threshold that is selected. Currently, the selection of this threshold is a manual procedure. Our work can act as a testbed for potential automatic threshold selection algorithms.

Additionally, our research can build upon recent work in image forgery detection using deep neural networks, and siamese neural networks in particular [15]. It would be interesting to see if the same neural network model trained on image forgery data would accurately detect fraudulent images. This particular methodology shows the potential to be more robust than ELA, and this is a promising direction for our research.

## X. CONCLUSION

In this paper, we proposed a system that addresses the problem of face verification and image forgery detection. We frame the problem as a one-shot learning task, and use data augmentation to enrich our dataset, and account for variations in pose. We also use image processing techniques to remove the background from the image. The face verification system consists of a training phase and an inference phase. We explored different loss functions to train our model, and these experiments show that the siamese neural network with triplet loss function yields the highest accuracy of 85% on Pre-post images. We also evaluate the importance of background removal, and found that this improves the accuracy of our model. Another parameter that we tested during training was whether a model trained on a specific combination of images is able to generalize on other image subsets.

During the inference phase, we first employ Error Level Analysis to detect image forgeries. After an image pair is found to be genuine, it is passed to the face verification system. The proposed system utilizes a deep convolutional network to convert the input image of a patient into a low dimensional embedding. This embedding is looked up in the database and verification is done if the L2 norm is below a threshold. Thus, our system can act as an alternative to manual verification of images, and it does so consuming minimal data.